# Pattern Based Term Extraction Using ACABIT System

Koichi Takeuchi[†]　　Kyo Kageura[†]　　Teruo Koyama[†]

Beatrice Daille[‡]　and　Laurent Romary[‡‡]

[†] National Institute of Informatics　　2-1-2 Hitotsubashi, Chiyoda-ku, Tokyo, 101-8430 Japan
[‡] University of Nantes　　1, quai de Tourville BP 13522, 44035 NANTES CEDEX 1, France
[‡‡] Loria　　Campus Scientifique, B.P. 239, 54506 VANDOEUVRE-lès-NANCY CEDEX, France

E-mail:　[†] {koichi,kyo,t_koyama}@nii.ac.jp,　[‡] Beatrice.Daille@irin.univ-nantes.fr,　[‡‡] Laurent.Romary@loria.fr

**Abstract**　In this paper, we proposed pattern based term extraction model for Japanese applying ACABIT system developed for French. Proposed model evaluates termhood using morphological patterns of basic terms and term variants. After extracting term selections, ACABIT system filters non-terms out from the selections based on simple log likely hood evaluation. This approach would be suitable to Japanese term extraction because most of Japanese terms form compound nouns or simple phrasal patterns. After showing the morphological patterns for terms, we show experimental results. By comparing morphological patterns with foreign languages, we discuss morphological units in Japanese.
**Keyword**　Term extraction, Pattern based, Morphological patterns, Termhood

## 文法パターンに基づく用語抽出モデルの構築

竹内　孔一[†]　　影浦　峡[‡]　　小山　照夫[‡]

[†] 国立情報学研究所　　〒101-8430　東京都千代田区一ツ橋 2-1-2
[‡] University of Nantes　　1, quai de Tourville BP 13522, 44035 NANTES CEDEX 1, France
[‡‡] Loria　　Campus Scientifique, B.P. 239, 54506 VANDOEUVRE-lès-NANCY CEDEX, France
[‡] R&D Division, Osaka Corporation　　4-5-6 Kawada, Suita-shi, Osaka, 565-0456 Japan

E-mail:　[†] {koichi,kyo,t_koyama}@nii.ac.jp,　[‡] Beatrice.Daille@irin.univ-nantes.fr,　[‡‡] Laurent.Romary@loria.fr

**あらまし**　文法パターンを利用した日本語の用語抽出モデルの構築を行う。用語抽出を行うためには用語らしさをいかに評価することが重要である。我々は用語らしさを「語の意味的構成の強さ」と「分野の特殊性」の2点に分解し、そのうち前者について用語抽出モデルの構築を行った。品詞パターンを利用して語彙的な結束性が強いパターンを記述し、フランス語の用語抽出モデルとして開発された ACABIT システムに適用し用語抽出実験を行った。この実験を通して日本語の品詞パターンと品詞パターンの元となる日本語の語彙の単位について多言語的観点から考察する。
**キーワード**　用語抽出, パターンベース, 文法的パターン, 用語らしさ

## 1. Introduction

In this article, we propose a pattern based term extraction model and show experimental results that it produces.

The difficulty of term extraction is how to evaluate "termhood'' for inputted word sequences. Since the role of term is to denote one special concept in some domain, termhood should be evaluated according to the following two aspects: the first one is the strength of unity for component words as a term, and the other one is the domain specialty of the word. Most of the previous (see Kageura et al. (2000)) approaches take the latter approach and focuse on the development of evaluation methods of domain specialty based on the statistics of words in documents. However, the former evaluation approach, that is, unity as a term is very important because the development of the approach is directly related to the mechanisms of creation of a new concept by composing words.

Recentl,y some term extraction works have focused on the unity of words. Nakagawa (2000) proposed a statistical method to evaluate the unity of words based on the strength of the connection between constituent words for Japanese terms. Jacquemin (1996) introduced the idea of term variants assuming complex terms must be derived from the basic terms because unity of term must be kept between basic term and its variants. He developed a pattern based extraction model for complex terms in English and French. Yoshikane el al. (2003) applied Jacquemin's approach to Japanese terms. Extending this approach, Daille (1996) proposed morpho-syntactic pattern based model (implemented in a system called ACABIT) to extract basic terms and term variants for French without basic term list.

In this paper, we propose a pattern based term extraction model for Japanese applying Daille's approach. It must be suitable to Japanese term extraction because most of Japanese terms form compound nouns then morpho- syntactic variation of pattern would be limited. The aim of this research is to identify the difference of mechanisms of creating new concepts as terms between French and Japanese by comparing morpho-syntactic patterns.

In the following sections, after presenting an overview of extraction systems and grammatical patterns for Japanese terms, we show some experimental results.

## 2. Approach
### 2.1. ACABIT

In the ACABIT system all complex terms are regarded as derivations from basic two word terms, the system tries to extract basic terms and term variants using morpho-syntactic patterns. The grammatical patterns are described based on POSes. The input of ACABIT is a POS-tagged text that is annotated by POS-tagger at the preprocessing stage(see Fig.1). From apractical point of view, the morpho- syntactic patterns should thus be designed according to the POS set of preprocessor. The output of the ACABIT system is a list of basic terms and term variants. Terms extracted by grammatical patterns are evaluated by log likelyhood that denotes the strength of connection between words. The outputs are convenient for the application of a deeper analysis because the relations between derived terms and basic terms are designated in the term lists.

ACABIT is basically designed for the extraction of French terms, but it can be applied to other languages by changing the POS-tagger and the morpho- syntactic patterns.

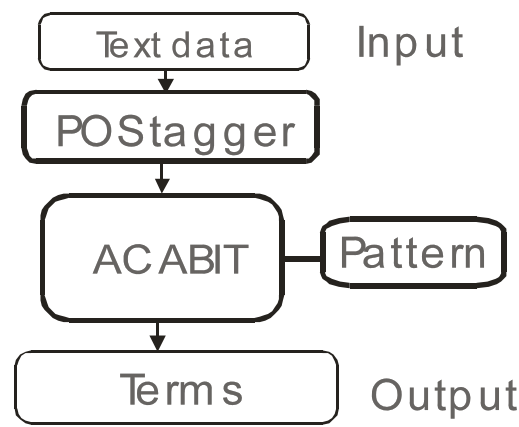

Fig. Overview of ACABIT system.

### 2.2. Overview of Japanese ACABIT

We construct a Japanese term extraction model applying ACABIT. As a POS-tagger, we selected ChaSen (Matsumoto et al. (1996)) that is a morphological analyzer for Japanese. It has about 100,000 word entries in the dictionary and 40 kinds of shallow syntactic tags as POSes. The morpho-syntactic patterns in Japanese ACABIT system are constructed based on the POS set of ChaSen.

### 2.3. Morphological patterns

In our Japanese term extraction model, we also assume that all complex terms would be derived from basic terms that compose compound nouns or noun phrases. In the following, after we show morphological patterns of basic words for Japanese terms, we show the patterns of complex terms derived from them.

### 2.3.1. Patterns for basic terms

In Japanese, a morphological head of intra-compounds or phrase is a final word. We design patterns for basic terms enough to be compound noun or noun phrase according to morphological characteristics.

**1. Noun-Noun:** This basic combination of compounds is the most popular one, but there is some specialty of Japanese. In this pattern, "Noun" does not mean only nominal noun, but also deverbal noun and deadjectival noun. Deverbal noun[1] acts basically as noun in sentence as itself but it can be a verb followed by the auxiliary verb *suru*, that is, it has both the characteristics of a noun and a verb. Similarly, a deadjectival noun acts as a noun in a sentence when on its own, but it can be an adjective followed by *na* inflection. Both deverbal nouns and deadjectival nouns are able to come to head as well as modifier position as nominal noun. Because of the existence of these two special nouns, long complex compounds often appear in Japanese sentence.

---
[1] It is the same thing that called *sahen-verb* in Yoshikane et al. (2003).

Taking for example,

*densi    kaigi*
electric conference
(electric conference)

consists of nominal nouns,

*johou    kensaku*
information   retrieve
(information retrieval)

is made of a nominal noun and a deverbal noun, and

*anzen    taisaku*
safe(-ty)   measure
(measure of safety)

is an adjectival noun followed by a deverbal noun.

In the following explanation, noun normally means nominal noun, deverbal noun and adjectival noun.

**2. Prefix-Noun:** The Prefix in Japanese forms one word, while usually prefixes are incorporated in words in European languages such as English and French. For example,

*fu    kanou*
im   possible
(impossible)

is a case when the word *fu* corresponds to the prefix "im" in English and French. The word *kanou* is a deadjectival noun. Depending on the meaning of the word, some prefixes correspond to nominal nouns.

*dai    youryou*
large   capacity
(large capacity)

**3. Noun-Suffix:** Suffixes in Japanese also form one word being different from suffixes in English and French, but they designats a function of composed word because head comes to final in Japanese word formation. For example, a suffix *hou* derives noun with deverbal noun:

*gousei    hou*
composite(-ion) method
(composition method),

and a suffix *ka* derives deverbal noun with noun or deadjectival noun:

*saiteki    ka*
optimal   -ize(-ion)
(optimization).

Comparing with English, suffix *ka* corresponds to two suffixes -ize and -ion.

**4. Noun-SuffixStem-Noun:** There is a special suffix that is very frequent and derives stem of adjectival noun from modifier noun. The pattern of basic term consists of three words.

*tetsuduki teki chishiki*
procedure –al knowledge
(procedural   knowledge)

Comparing with the translation in English, the function of the suffix *teki* is to make an adjectival modifier become a noun in the compound.

**5. Vinf-Noun:** Vinf denotes a verb with inflection except deverbal noun.[2] The inflection type depends on the meaning of the term. The example of passive voice is as follows:

*toji-ta    hairetsu*
close-ed   array
(closed array).

The inflection *ta* denotes the passive voice of the root verb *toji* (close). The example of active voice is as follows:

*yurag-i    zatsuon*
fluctuate-INF   noise
(fluctuation noise)

**6. VInf-Suffix:** Nominal suffix as pattern **3** can also take inflected verb that is except deverbal noun. Example is as follows.

*uketor-i    gawa*
receive-INF   side
(receiving area)

The "suffixstem" that derives stem of adjectival noun does not compose this form. As a whole, this multiword forms the middle between compound noun and phrase.

**7. AInf-Noun:** AInf means an adjective with inflection except deadjectival noun.[3] An example is as follows.

*fuka-i    chishiki*
deep-INF   knowledge

---

[2] These kinds of verb originated in traditional Japanese, while deverbal noun in China. They called *kango* and *wago*, respectively (see Section 2.3.2)

[3] This adjective originated in traditional Japanese words that is called *wago* (see Section 2.3.2).

(deep knowledge)

**8. Adj-SuffixNominalize-Noun:** There is a suffix that derives noun from adjectives that is the same as previous pattern.

<div style="text-align:center">
naga-sa    zokusei<br>
length    attribute<br>
(length    attribute)
</div>

In patterns **7** and **8**, inflection '-i' and suffix '-sa' can connect to all adjectives, for example *fuka-i* (deep) and *fuka-sa* (depth). If we now compare with English, the Japanese suffix '-sa' seems to be one inflection type of adjective. This grammar set comes from ChaSen as well as standard school grammar in Japanese, but this kind of gap would be regarded through comparing morphemes with a foreign language.

The above is the complete set of patterns for basic terms. From a syntactical view of composition, the patterns from **1** to **4** and **8** form compound nouns, while the patterns from **5** is phrase and **6** and **7** are middle. We have to take care of these type differences because there exist some limitations of composing term variants from these basic terms depending on the differences (see the next section).

### 2.3.2. Patterns for complex terms

Complex terms may form a compound noun or a phrase that is derived form basic terms. Because of specialty of Japanese words, composition types of compounds are limited according to the characteristics of word types: the one is "imported word" (IW) that originates in foreign language and the other is "word originating in traditional Japanese" (WJ). In compounding, there is a tendency to connect only the same types of words, especially, words in IW group can make long compound nouns with less difficulty, while WJ does not. We show this phenomenon in the following section. Besides the types of phrasal patterns as terms are also limited because most of terms form compound nouns (see Section 3). This is very different from English or French.

#### 2.3.2.1. Compound nouns

**Element+:** terms form compounds.
  **Element:** Noun-Noun, Prefix-Noun, Noun-Suffix, Number-Suffix, Number and Symbol.

Here Number means a character sequence of numbers, and Symbol is that of symbols. The pattern **Element+** denotes continuing more than one Element that forms compound nouns. This long compound noun is a special characteristic of basic terms in Element that consists of IWs. Adding basic terms of Noun-Noun, Prefix-Noun and Noun-Suffix, Number-Suffix, Number and Symbol are involved as Element. These tree types usually need another element to be terms, therefore they are Element but not basic term patterns. The example of "Number Suffix Noun" is as follows:

<div style="text-align:center">
2    sou    haisen<br>
two    layer    wiring<br>
(two-layer wiring).
</div>

Theoretically there is no limitation of connection, but practically we set the limitation of maximum length as 9.

#### 2.3.2.2. Limited patterns of term variants

The basic terms that contain WJ also form compound nouns or phrases, however, variation patterns of the basic terms are limited. This limitation comes from termhood because there is a preference to select compounds or some short phrasal expressions much more than syntactic long phrase so that one term expresses one static concept. In the following, we will show the limited patterns of terms variants for basic terms containing WJ and examples.

  **Adj-SuffNominalize-Noun+:**

<div style="text-align:center">
fuka-sa    yuusenn    tansaku<br>
deep-th    first    search<br>
(depth first search)
</div>

 **Noun-Vinf-Suffix:**    bitto ayamar-i      ritsu
<div style="text-align:center">
bit    make error-INF    rate<br>
(bit error rate)
</div>

 **Noun-Vinf-Noun:**    kyoushi    ar-i    gakusyuu
<div style="text-align:center">
supervisor    exist-INF    learning<br>
(supervised learning)
</div>

In the later two cases, nouns at the modifier position of Vinf are argument of the root verb (ex. *Kyoshi* (Supervisor) is an argument of *ar-i* (exist).).

Next case is complex term variants consists of IWs.

  **Prefix-Noun-Suffix-Noun*:**

<div style="text-align:center">
hi    douki    shiki<br>
not    synchronous    method<br>
(asynchronous system)
</div>

"Noun*" denotes 0 or more than 1 word of Noun is going to continue. The example is the case of **Prefix-Noun-**

**Suffix.** Of course Prefix-Noun-Noun-Suffix case has already been taken into account at **Element+** pattern.

### 2.3.2.3. Phrase (Syntactical compounding)

**(Element+)-of-(Element+)**

moji    no    daishou    jynjyo
character of   large-small   order
(collating sequence)

Theoretically, all compound nouns and noun phrases can be applied to this phrasal pattern. Besides this A of B phrase can be extended recursively as A of B of C. However we only permit the pattern of "A *no* (of) B" as variants for complex terms because the meaning of relation *no* is ambiguous and then we hesitate about using long sequence like "A *no* B *no* C" as a term.

### 2.4. How to apply patterns

We implement Japanese morphological patterns into ACABIT system. The important point how to apply them is order of patterns. We take the strategy to apply them from longer patterns so that system does not decompose long sequence term into short ones.

## 3. Experiment and results

We have two types of experiments in order to evaluate performance of Japanese ACABIT system. The fist evaluation is about the coverage of morphological patterns we elaborated. We input technical terms to ACABIT and check the rate of acceptability. The second is to evaluate term extraction performance. Since we do not have the all term set to some domain, we can only evaluate precision of our ACABIT. In order to do this experiment, we use the set of abstracts and author's keys that are distributed by NII for term extraction competition (Kageura 2000).

### 3.1. Coverage experiment

We prepare three kinds of technical terms: 1) technical term dictionary of information processing (ipdic) (Aiiso 1996), 2) term dictionary in computer domain (comdic) (Nichigai 1998) and 3) author's keywords in artificial intelligent domain (Kageura et al. 2000).

All terms are analyzed using ChaSen so that all terms are decomposed into basic word with POS.[4] After this process, we evaluate statistical characteristics of terms about: number of one word terms and number of phrasal terms.

Table 1 Statistics of input terms

|  | ipdic (%) | comdic (%) | jsai (%) |
|---|---|---|---|
| One word term | 2207/16275 (13.6) | 4480/38785 (11.6) | 658/4206 (15.6) |
| Phrasal term | 409/16275 (2.5) | 2366/38785 (6.1) | 231/4206 (5.5) |

From Table 1, the share of one word term is not a little, that is, over than 10 % for every source. In our approach, ACABIT does not extract one word terms because we assume that all terms consist of more than one word. The rates are upper bound of extraction. While phrasal terms are very few for every source, so our morphological patterns would be work well.

Table 2 shows the results of coverage performance of Japanese ACABIT.

Table 2 Coverage of Japanese ACABIT

|  | ipdic (%) | comdic (%) | jsai (%) |
|---|---|---|---|
| coverage | 4080/16275 (74.9) | 10623/38785 (72.6) | 3056/4206 (72.7) |

Our ACABIT works well although the upper bound of this experiment is about from 86 to 89 % for these terms. The error types are categorized into three: **a)** variety of term, **b)** error of annotation of ChaSen, and **c)** lack of pattern of ACABIT. We explain them as follows.

**a)** Most errors occur on terms that contain proper nouns. For example, *nyuuton* (Newton) in *nyutonn hou* (Newton method) is annotated proper noun in ChaSen. This is correct analysis for general purpose while proper noun would not be a term. There is a possibility to make some patterns based on suffixes such as "ProperN Suffix (method)", but it cannot cover another variety such as "Microsoft network".

**b)** ChaSen makes annotation errors on ambiguous words that are, for example, *douki* that is deverbal noun is miss annotated as adverbial noun. This problem relates lack of study of Japanese adverbial words.

**c)** ACABIT system makes errors because of lack of patterns. Most of them are phrasal expression but are lexicalized such as *1 no hosuu* (one's complement). We cannot make a pattern like Number-of-Noun to accept this because the pattern usually means phrase.

---
[4] Basic performance of morphological analyzer Chasen is over than 95%.

## 3.2. Term extraction

Japanese ACABIT is applied to abstract texts in artificial intelligence domain in order to show term extraction performance of ACABIT. Assuming atuthor's key words are correct terms for the texts, we evaluate the performance of ACABIT by comparing extracted terms with author's key.

Table 3 shows the statistics of author's key. According to the table, 68.7% keys are involved in abstract text and 20.1% of them are one-word keys, then 2308 words are the upper bound keys to be extracted.

Table 3　Statistics of author's key

|  | author's key (%) |
|---|---|
| contained in text | 2890/4206 (68.7) |
| one word key | 582/2890 (20.1) |
| Upper bound | 2308/4206 (54.9) |

Table 4 shows the results of term extraction comparing with author's key. words.[5] We evaluate precision, correct rate to all author's key and correct rate to upper bound of author's key.

Table 4　Results of term extraction

|  | jsai author's key |
|---|---|
| precision | 1639/ 23494 (7.0) |
| hit rate to all keys | 1639/4206 (39.0) |
| hit rate to upper bound | 1639/2308 (71.0) |

All extracted terms are evaluated. In Table 4, our ACABIT works well because 71% of upper bound keys are successfully extracted. The precision is, however, poor. Even when we filter out the low log likely value terms of ACABIT, precision was 20.8% and rate to upper bound was 25.0%.

Example correctly extracted words are *iden-teki-arugorizumu* (generic algorithm), *chishiki-beisu* (knowledge base), and wrong examples are *hon-ronbun* (this paper), *hon-kenkyu* (this research) *hissya-ra* (authors). The words that are extracted wrongly are high frequency words in target text.

---

[5] This is the rate such as recall but not recall, exactly because we do not know the correct all terms in this domain.

## 4. Discussion

From the experimental results of Section 3.1 and 3.2, we found that our morpho-syntactic patterns have good coverage for technical terms. However precision of term extraction is poor because we only apply simple log likely hood evaluation of ACABIT, at the moment. We will be able to improve the precision rates by applying more sophisticated statistical approach to evaluate the unity of intra-term structure as Nakagawa (2000).

Comparing with foreign language such as English and French, Japanese terms prefer to form compound noun while English and French prefer to form phrase about complex terms. In Japanese term extraction, we have to discriminate terms from general words on compound nouns while in European language on phrasal expressions.

## 5. Conclusion

We proposed pattern based term extraction method and show the experimental results. We try to extract terms using termhood, especially, focusing on the unity of intra-term structure by morpho-syntactic patterns. From the experimental results, our constructed patterns work well for coverage of terms but precision is not so grate. That problem is not the issue of this paper because that is a problem how to evaluate specialty of words in some domain. Comparing morphological patterns in English and French, we clarify the difference of composing level for terms: most of Japanese terms form compound noun while English and French terms form phrasal patterns.